\documentclass[a4paper,UKenglish,cleveref, autoref, thm-restate]{tgdk-v2021}

\usepackage{booktabs}
\usepackage[section]{placeins}

\title{Hierarchical Retrieval with Out-Of-Vocabulary Queries: A Case Study on SNOMED CT}

\titlerunning{HR with OOV Queries}

\author{Jonathon {Dilworth}}{The University of Manchester, United Kingdom \and \url{http://www.dilworth.io}}{jon@dilworth.dev}{https://orcid.org/0009-0009-0260-0492}{}
 
\author{Hui {Yang}}{The University of Manchester, United Kingdom \and \url{}}{hui.yang-2@manchester.ac.uk}{https://orcid.org/0000-0002-4262-4001}{}

\author{Jiaoyan {Chen}}{The University of Manchester, United Kingdom \and \url{https://chenjiaoyan.github.io/}}{jiaoyan.chen@manchester.ac.uk}{https://orcid.org/0000-0003-4643-6750}{}

\author{Yongsheng {Gao}}{SNOMED International, United Kingdom \and \url{}}{yga@snomed.org}{https://orcid.org/0000-0002-3468-2930}{}

\author{Ernesto {Jiménez-Ruiz}}{City St George's, University of London, United Kingdom \and \url{https://ernestojimenezruiz.github.io/}}{Ernesto.Jimenez-Ruiz@citystgeorges.ac.uk}{https://orcid.org/0000-0002-9083-4599}{}

\authorrunning{J. Dilworth, H. Yang, J. Chen, Y. Gao, and E. Jiménez-Ruiz} 

\Copyright{Jonathon Dilworth, Hui Yang, Jiaoyan Chen, Yongsheng Gao, and Ernest Jiménez-Ruiz} 

\ccsdesc[500]{Computing methodologies~Knowledge representation and reasoning}
\ccsdesc[500]{Computing methodologies~Natural language processing}

\keywords{Ontology Embedding, Language Model, Hierarchical Retrieval, Out-Of-Vocabulary Query, SNOMED CT}

\category{} 

\relatedversiondetails[cite={dilworth2025hierarchicalretrievaloutofvocabularyqueries}]{Prior Versions}{https://arxiv.org/abs/2511.16698} 

\supplement{GitHub}\supplementdetails[]{GitHub}{https://github.com/jonathondilworth/HR-OOV-SNOMED-CT}

\acknowledgements{This work is funded by the EPSRC project OntoEm (EP/Y017706/1).}

\nolinenumbers 

\Volume{1}
\Issue{1}
\Article{42}
\DateSubmission{Date of submission}
\DateAcceptance{Date of acceptance}
\DatePublished{Date of publishing}
\SectionAreaEditor{TGDK section area editor}

\begin{document}

\maketitle

\begin{abstract}
SNOMED CT is a biomedical ontology with a hierarchical representation, modelling terminological concepts at a large scale. Knowledge retrieval in SNOMED CT is critical for its application but often proves challenging due to linguistic ambiguity, synonymy, polysemy, and so on. This problem is exacerbated when the queries are out-of-vocabulary (OOV), i.e., lacking any equivalent matches in the ontology. In this work, we focus on the problem of hierarchical concept retrieval from SNOMED CT with OOV queries, and propose an approach driven by utilising language model-based ontology embeddings, which represent hierarchical concepts in a hyperbolic space for enabling efficient subsumption inference between a textual query and an arbitrary concept. For evaluation, we construct three datasets where OOV queries are annotated against SNOMED CT concepts, testing the retrieval of the most specific subsumers and their less relevant ancestors. We find that our method outperforms the baselines, including SBERT, SapBERT, and two lexical matching methods. While evaluated against SNOMED CT, the approach is generalisable and can be extended to other ontologies. 
We release all the experiment codes and datasets at \url{https://github.com/jonathondilworth/HR-OOV-SNOMED-CT}.
\end{abstract}

\section{Introduction}
\label{sec:introduction}

OWL (Web Ontology Language) ontologies are formal, machine-interpretable and shared representations of knowledge with explicit semantics \cite{OWL2primer}. These knowledge representations are modelled hierarchically through subsumption between concepts as well as expressions in Description Logic. Their usefulness is evidenced by wide adoption in industries such as healthcare, where high-quality, structured domain knowledge can aid in decision support, clinical reporting, and biomedical research. Importantly, the use of ontologies in the healthcare domain is already well established, with semantic interoperability between clinical systems often being ontology-driven. For instance, SNOMED CT is a terminological ontology that supports healthcare information systems in this manner \cite{el2018snomed,chang2021use}. Given the reliance on such terminologies, effective retrieval is a key consideration in improving usability \cite{bakhshi2012usability}.

\begin{figure}[htb!]
  \centering
  \includegraphics[width=0.7\columnwidth]{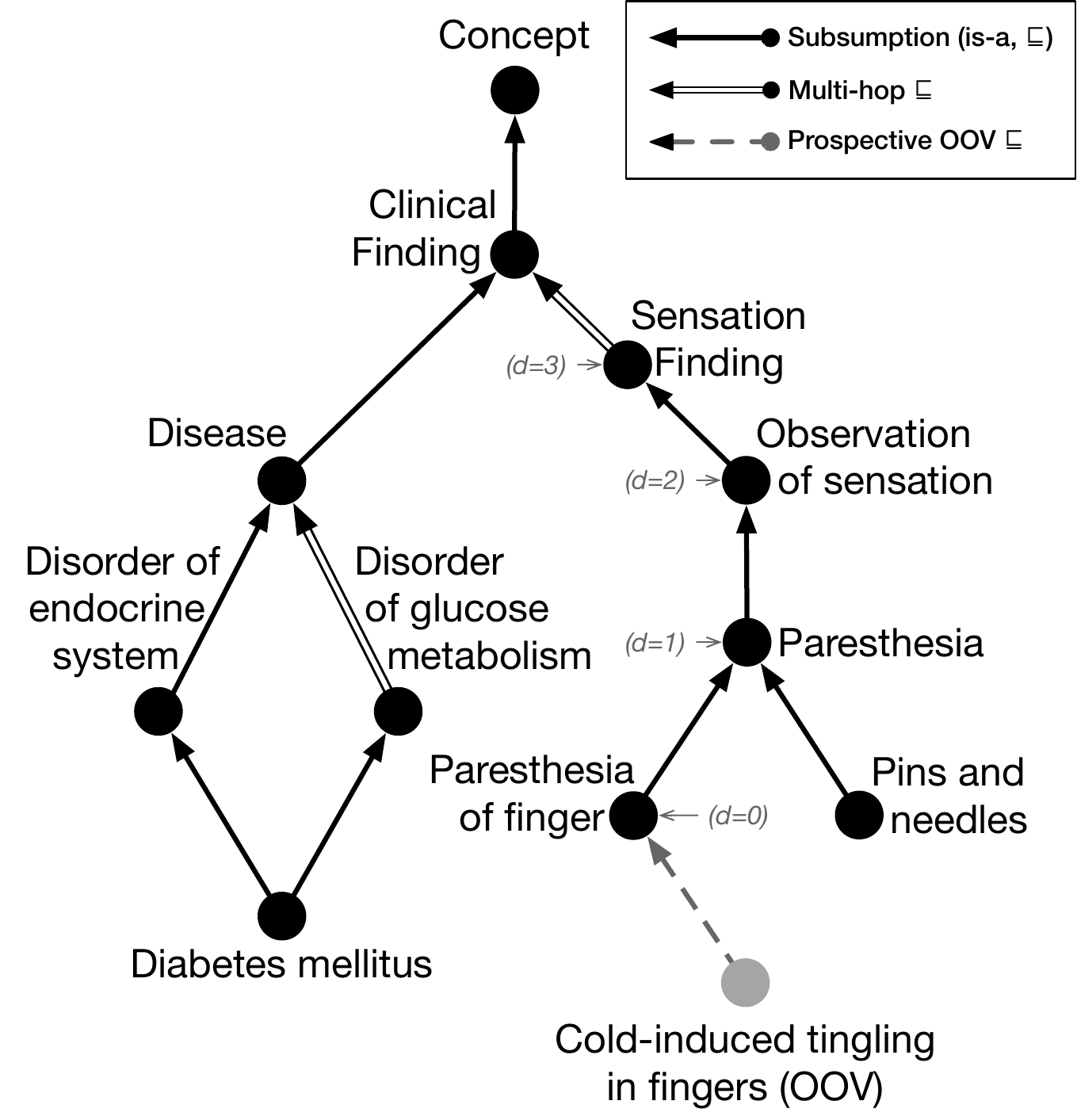}
  \caption{A concept hierarchy fragment from SNOMED CT which is represented by black nodes and solid arrows, and an OOV query which is represented by a gray node. The dashed grey arrow shows the prospective subsumption relationship between the OOV query ``Cold-induced tingling in fingers'' and its most specific (and direct) subsumer \textit{Paresthesia of finger}.
  The depth $d$ is the number of direct subsumption hops from the most specific subsumer of the query to a concept.
  Note that the double-stroke arrows represent multi-hop subsumption (i.e., some intermediate concepts and subsumptions have been ommited for visualisation).
  }
  \label{fig:hierarchy-fragment}
\end{figure}

Retrieval in SNOMED CT is often implemented through lexical matching, such as in the SNOMED CT browser\footnote{https://termbrowser.nhs.uk/?perspective=full}. These methods rely on matching exact keywords or phrases from the search input against descriptions of concepts in SNOMED CT. Meanwhile, embedding-based methods, such as SBERT \cite{reimers_sentence-bert_2019}, have also been widely studied in retrieving concepts with similar semantic meanings. Some ontology embeddings such as OWL2Vec* extend embedding-based approaches by training on the ontology's own contents \cite{chen_owl2vec_2021,chen2025ontology}, thereby being able to improve ontology-specific retrieval.
However, these lexical matching methods rely on surface-form overlap, and the text-aware embedding methods can only capture and represent equivalence through vector similarity. Both approaches may struggle to handle search terms that have no equivalently matched counterparts in the ontology, i.e., out-of-vocabulary (OOV) queries. 
For example, in Fig.~\ref{fig:hierarchy-fragment}, the query ``Cold-induced tingling in fingers'' has no equivalent matches in SNOMED CT, but its intended clinical concept is directly subsumed by the concept \textit{Paresthesia of finger}\footnote{Paresthesia is a tingling or numbness sensation, often likened to ``Pins and needles''.} and this subsumer can be returned as a useful search result. 

Such OOV queries are common in SNOMED CT retrieval.
Based on an analysis of 54986 real-world search queries toward the SNOMED CT browser from 07/12/25 to 11/12/25, we find only 15.1\% of them have SNOMED CT concepts with lexically matched terms.

This data analysis result is consistent with many scenarios of querying SNOMED CT. For example, when patients visit a doctor or search through a healthcare information system, they lack knowledge of professional clinical terminology; thus, the terms used to describe diseases and symptoms are often close but not equal to the clinical terms. Even doctors are often unable to search with exact clinical terms, considering the large scale, varying levels of specificity, and high complexity of clinical terminology.

To return promising results for these ontology OOV queries, we present a new method of applying semantic embedding for ontology hierarchical retrieval (HR), where the results for a natural language searching keyword or phrase are defined as its parent and ancestor concepts drawn from the ontology's hierarchical concept structure \cite{you2025hierarchical}. In particular, two language model-based ontology embedding methods are applied: the Hierarchy Transformer (HiT) \cite{he_language_2024}, which aims to embed a concept hierarchy (i.e., a taxonomy), and the Ontology Transformer (OnT) \cite{yang2025language}, which can embed an ontology's concept hierarchy, existential restrictions and concept conjunctions. They both encode concept labels via re-training a pre-trained encoder-based language model and efficiently preserve the concepts' hierarchy in a hyperbolic space. The trained embedding model allows for direct subsumption inference between concepts indicated by two arbitrary phrases by calling a depth-based scoring function.

In this work, we propose two settings for ontology HR with OOV queries: \textit{optimal target}, where the most specific subsumer is expected to be returned, and \textit{multi-hop target}, where any valid subsumer within a defined distance is expected to be returned.
For the example OOV query in Fig.~\ref{fig:hierarchy-fragment}, \textit{Paresthesia of finger} is both an optimal target and a multi-hop target, while \textit{Paresthesia} is a multi-hop target but not an optimal target.
We construct three datasets for evaluation.
The first--EVAL-100 is a set of queries which are lexically disjoint from all SNOMED CT concept descriptions, including synonyms. We construct this dataset by extracting named entities from the MIRAGE benchmark \cite{xiong-etal-2024-benchmarking}, then manually reviewing and annotating them with target concepts from SNOMED CT. The second and third datasets--OET-CPP and OET-Disease--are both derived from the existing OET datasets, which are for benchmarking methods for the task of enriching an ontology by placing new concepts extracted from text corpora \cite{dong2023ontology}. 
The experiments demonstrate that our ontology embedding-based HR method outperforms baselines using lexical indexing and Sentence-BERT embedding in both optimal and multi-hop target settings, with OnT performing similarly to HiT in both settings. While this work has a particular focus on SNOMED CT, the approach is generalisable and can be applied to other ontologies, improving their accessibility and usability in downstream applications like clinical decision support and terminology navigation. 

The contributions of this work can be summarised as follows:
\begin{enumerate}
    \item We investigate the task of HR over an ontology with OOV queries with a case study on SNOMED CT. In particular, we define the setting of optimal target, which is expected to retrieve the direct subsumers of the query, and the setting of multi-hop target, which allows to return subsumers within a specific number of hops in the hierarchy structure.
    \item We propose a novel ontology embedding-based framework for addressing the above task of HR with OOV queries. Specifically, we implement the framework using two language model-based ontology embedding models --- the Hierarchy Transformer (HiT) and Ontology Transformer (OnT), both of which efficiently preserve concept hierarchies in a hyperbolic space and can infer the subsumption between two concepts with their labels.
    \item We construct three datasets for the HR task over SNOMED CT based on manual annotation and the existing concept placement benchmark. Extensive evaluation on these three datasets can verify the effectiveness of our framework, which outperforms different lexical matching and embedding-based retrieval methods.
    For example, on average, it has 16 higher mean reciprocal rank (MRR) than the best baseline in the deepest multi-hop target setting.
\end{enumerate}

The remainder of this paper is structured as follows. In Section~\ref{sec:related-work}, we survey related work, reviewing existing retrieval methods for SNOMED~CT. In Section~\ref{sec:preliminaries}, we introduce the preliminaries:  SNOMED~CT, the hyperbolic geometry, HiT and OnT. The methodology is described in Section~\ref{sec:task-definition-and-methodology}, alongside a formal problem definition. In Section~\ref{sec:datasets}, we introduce how our datasets are constructed. The experimental setup, results and discussion are then provided under Section~\ref{sec:experiments}. Finally, we conclude this work and introduce the future work in Section~\ref{sec:conclusion}.

\section{Related Work}
\label{sec:related-work}

Traditional ontology retrieval systems like the SNOMED CT browser often use a combination of lexical matching techniques such as TF-IDF and BM25 over different textual annotations of the concepts, and are usually implemented based on an inverted index \cite{manning2008introduction,robertson2009probabilistic}. 
While these systems often provide promising results, the lexical matching they rely on focuses on the surface form and often fails to capture and compare the underlying semantic meaning of terms. Consequently, they cannot handle natural language ambiguity and may achieve poor performance for out-of-vocabulary queries which are not lexically matched with any ontology concepts. 

To overcome the limitations of lexical matching, alternative methods based on word embeddings, such as Word2Vec \cite{mikolov2013distributed} and GloVe \cite{pennington2014glove}, have been proposed for measuring semantic similarity between phrases. 
These word embedding models can be further tailored to an ontology by re-training for more accurate similarity measurement, with typical examples of OPA2Vec \cite{smaili2019opa2vec} and OWL2Vec* \cite{chen_owl2vec_2021}.
Such early stage word embedding models are non-contextual, which means a word has one static vector no matter where it appears, and they are outperformed by the more recent contextual embedding models that are based on Transformer architectures. 
One classic contextual embedding model is BERT (Bidirectional Encoder Representations from Transformers) \cite{devlin2019bert}, which embeds a word conditioned on its surrounding words in the sentence it appears.
Sentence-BERT (SBERT) \cite{reimers_sentence-bert_2019} further extends BERT by a Siamese bi-encoder architecture for embedding phrases and sentences, and has achieved promising results calculating similarity between text. 
Such pre-trained Transformer-based encoders, also known as pre-trained language models, have been applied to concept retrieval in SNOMED CT, often in combination with different tasks like medical entity linking (a.k.a. clinic coding or concept normalisation), which is to match a mention in the text to an equivalent concept in an ontology \cite{chang2024use,kulyabin2024snobert}. 
For instance, SapBERT \cite{liu2021self}, which uses a self-alignment pre-training objective to train a Transformer-based encoder for embedding biomedical concepts, enables efficient and accurate entity linking via a nearest-neighbour search over candidate concept embeddings. 
Although there have been several works that apply BERT-based methods for SNOMED CT concept retrieval, they model the mention-to-concept mapping as an equivalence relationship. Such approaches of applying BERT alike embedding models will miss many directly relevant concepts in handling HR with OOV queries. 
As discussed in \cite{nawroth2021supporting}, ontology querying with \textit{emerging entities}, which are referred to as an \textit{``Anomalous State of Knowledge’’} and are very close to OOV queries, is a genuine open problem, and directly applying word embedding models cannot deal with the out-of-vocabulary challenge, while some additional, ad-hoc processing over the context is required.

In particular, some entity linking-based works including BLINKOut \cite{dong2023reveal} for linking mentions to NIL indicating there are no matched concepts and \cite{dong2024language, dong2023ontology} for inserting mentions into an ontology as new concepts have similar SNOMED CT concept retrieval scenarios as this work.  
The difference lies in the following aspects: (1) they consider mentions within a textual context like a sentence while we focus on queries which are usually isolated phrases; (2) their retrieval methods still return similar concepts with an apprxomiation of equivalence using fine-tuned BERT alike models, and then they search in the context of these concepts, while we expect to directly return subsumers through ontology embeddings models that preserve concept hierarchies. Actually, our HR method can be applied to these works for augmentation as a technical foundation of concept retrieval.

You et al. \cite{you2025hierarchical} propose to train a dual encoder model to deal with the problem of HR, where they assume a document set has a hidden hierarchical structure, and such a hierarchical structure should be learned in the encoder and utilised for retrieval. 
However, their model should be trained with a set of annotated samples (i.e., query-document pairs), while our method only relies on ontology embeddings that are trained on the ontology itself without using any query annotations.

\section{Preliminaries}
\label{sec:preliminaries}

\subsection{SNOMED CT}
\label{sec:snomed-ct}

SNOMED CT is an OWL ontology that contains entities which include concepts (classes) and roles (properties), and axioms that represent logical relationships between entities in Description Logic \cite{OWL2primer}. Concepts are organised through subsumption of the form $C \sqsubseteq D$, such as \textit{Diabetes Mellitus} $\sqsubseteq$ \textit{Disease}, and \textit{Disease} $\sqsubseteq$ \textit{Clinical Finding}. These subsumptions form a large-scale tree-like structure, known as the concept hierarchy (see example in Fig.~\ref{fig:hierarchy-fragment}). It supports inheritance reasoning via transitivity, enabling new subsumption inference.

Note that concepts can also be complex logical expressions\footnote{SNOMED CT is authored in OWL 2 EL, corresponding to the $\mathcal{EL}^{++}$ fragment of Description Logic; its logical expressions support conjunction and existential restriction, but exclude disjunction and universal restriction.} that are constructed with at least one logical operator. For instance, \textit{Diabetes mellitus} has two parents: \textit{Disorder of Glucose Metabolism} and \textit{Disorder of Endocrine System}. The latter is equivalent to a complex concept $\mathit{Disease} \sqcap \exists\mathit{FindingSite}.\mathit{StructureOfEndocrineSystem}$, which means a \textit{Disease} that is \textit{restricted to} a particular \textit{finding site}.

\subsection{Hyperbolic Space}
\label{sec:hyperbolic-space} 

A $d$-dimensional Riemannian manifold $\mathcal{M}$ \cite{lee2006riemannian} is defined as a smooth differentiable manifold equipped with a Riemannian metric tensor $g$. 
Hyperbolic space $\mathbb{H}^n$ is a Riemannian manifold with a constant negative sectional curvature $-\kappa$, which can be represented in the Poincaré ball model whose points lie within the open ball, given by:
 
\begin{equation}
B_{\kappa}^n = \{\ x \in \mathbb{R}^n : \|x\| < r\ \}, \qquad r=\frac{1}{\sqrt{\kappa}},
\end{equation}

\noindent where $r$ is the radius of the ball. The Poincaré metric $g_{\kappa}$ induces the geodesic distance function $d_{\kappa}$ between any two points $x,y \in B^n_{\kappa}$. This function is applied for scoring in Section~\ref{sec:task-definition-and-methodology} and is given by:

\begin{equation}
\label{eq:hyp-dist}
d_{\kappa}(x,y) = \frac{1}{\sqrt{\kappa}} 
\cdot 
\operatorname{arcosh} 
\Biggl( 1 + \frac{2\kappa \|x - y\|^{2}}{
  \bigl( 1 - \kappa \|x\|^{2} \bigr) \cdot \bigl( 1 - \kappa \|y\|^{2} \bigr)
}
\Biggr).
\end{equation}

\noindent As $\|x\|$ and $\|y\|$ approach the boundary of the ball (norm $\to \frac{1}{\sqrt{\kappa}}$), distances diverge even if the Euclidean norm difference $\|x-y\|$ is not itself significant, meaning that points situated near the boundary can represent more specific concepts (since their hyperbolic separation becomes large). This is in contrast to points situated toward the centre, which represent more generic concepts.

\subsection{The Hierarchy \& Ontology Transformer} \label{sec:hit-and-ont} 
The Hierarchy Transformer (HiT) \cite{he_language_2024} encode the hirerachy of concept by combining the encoder-based pre-trained language model SBERT \cite{reimers_sentence-bert_2019} with hyperbolic space embeddings (i.e., embeddings by the final layer output are situated within the Poincaré ball $B_{\kappa}^n$). In this architecture, SBERT captures the textual semantics, while the latter encodes the hierarchy of ontology concepts by leveraging the geometric properties of hyperbolic space.

To re-train SBERT for such embeddings, HiT uses a loss function compose of two parts: a hyperbolic clustering loss and a hyperbolic centripetal loss. Sepecifically, their training samples consist of triplets of the form $(x, x^+, x^-)$, where $x$ is a given named concept in the ontology, $x^+$ is a direct parent of $x$ defined in the ontology, and $x^-$ is a concept randomly selected from the ontology or from the sibilings of $x$.
$x$ and $x^+$ form a positive sample, while $x$ and $x^-$ form a negative sample. We use their bold forms $\boldsymbol{x}$, $\boldsymbol{x}^+$ and $\boldsymbol{x}^-$ to denote their embeddings output by the language model.  

\begin{figure}[!t]
  \centering
  \begin{subfigure}[b]{0.32\linewidth}
    \centering
    \includegraphics[width=\linewidth]{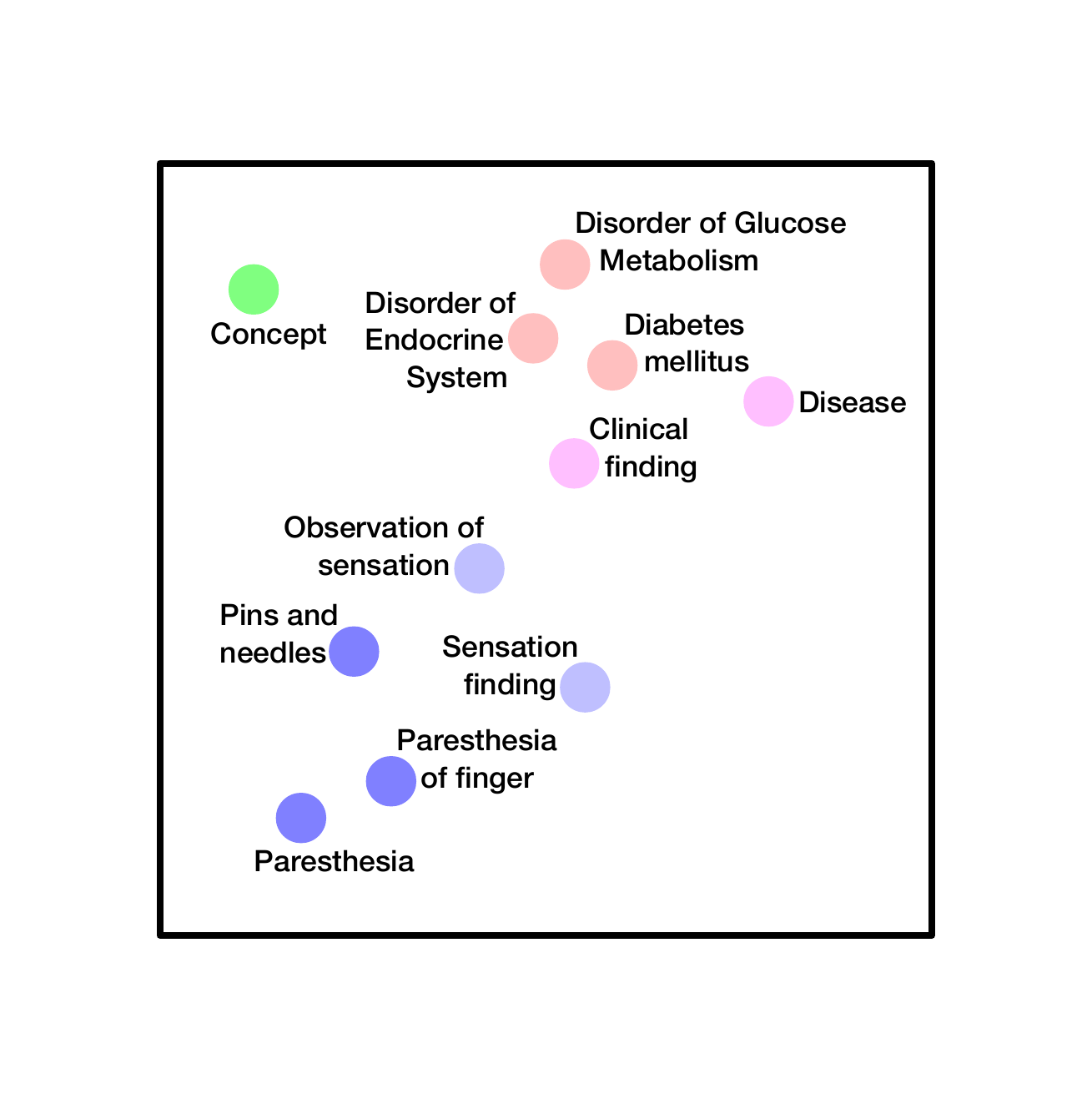}
    \caption{}
    \label{fig:hit-embedding-intuition-a}
  \end{subfigure}
  \hfill
  \begin{subfigure}[b]{0.32\linewidth}
    \centering
    \includegraphics[width=\linewidth]{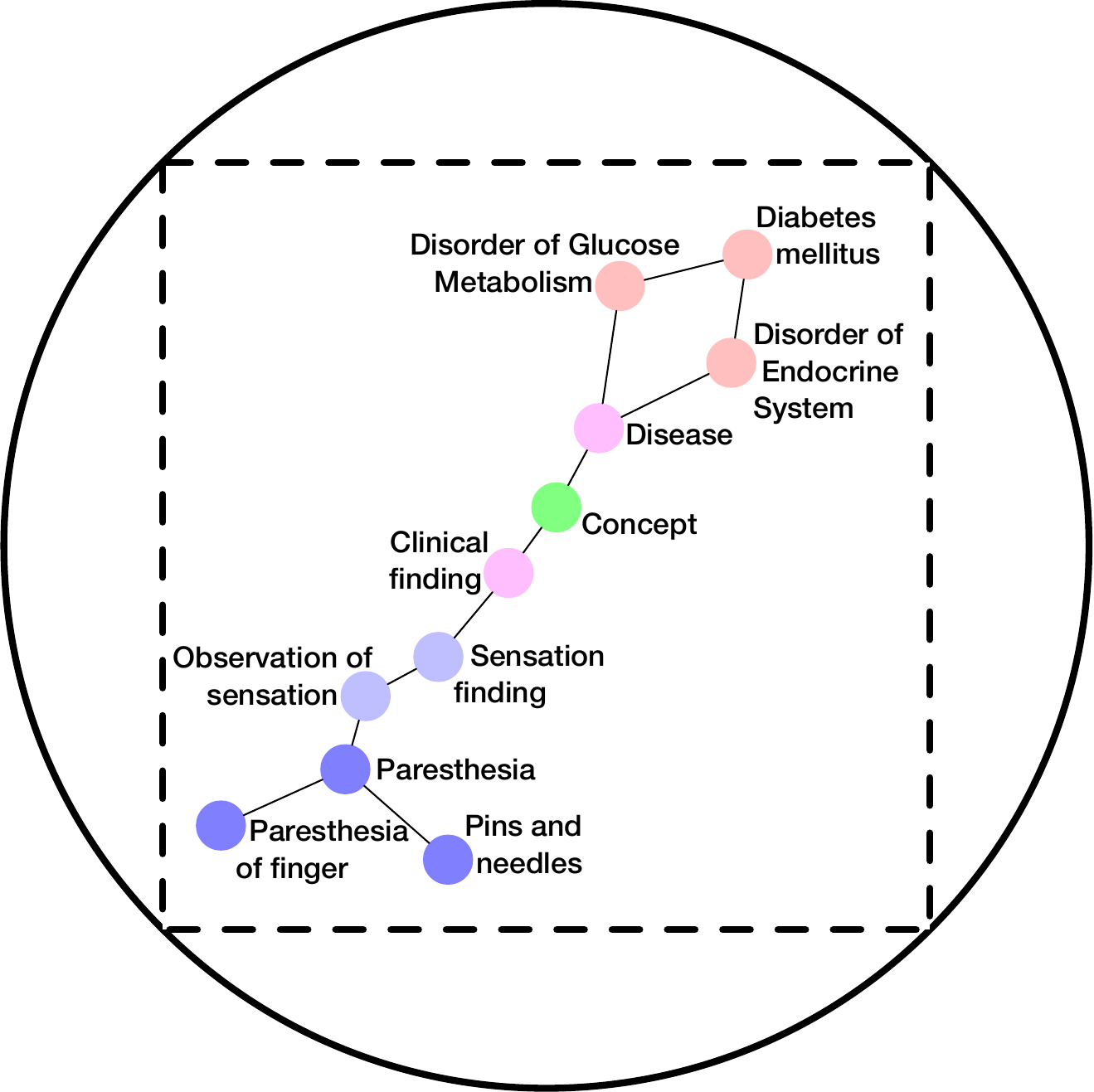}
    \caption{}
    \label{fig:hit-embedding-intuition-b}
  \end{subfigure}
  \hfill
  \begin{subfigure}[b]{0.32\linewidth}
    \centering
    \includegraphics[width=\linewidth]{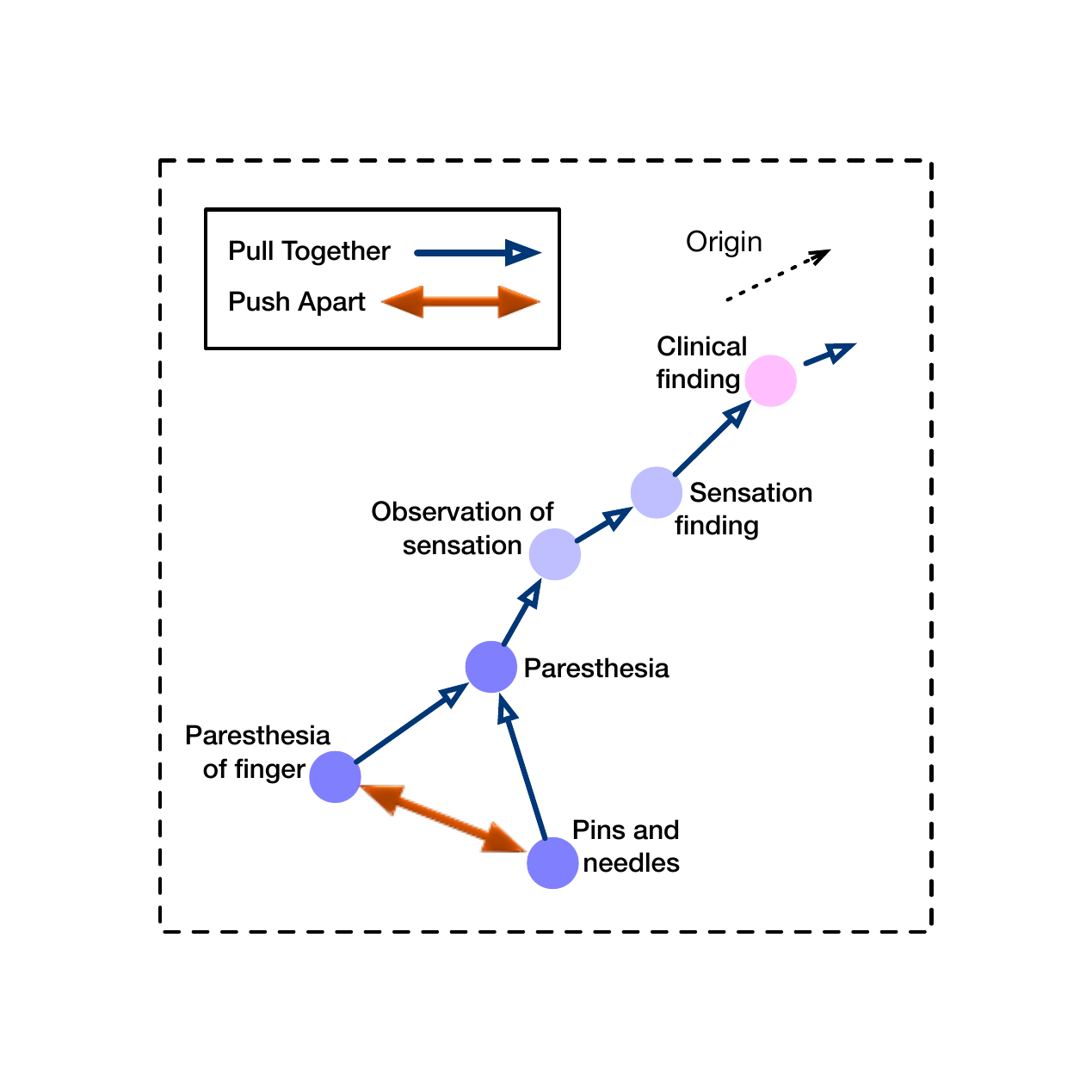}
    \caption{}
    \label{fig:hit-embedding-intuition-c}
  \end{subfigure}
  \caption{Illustration of HiT using concept branches in SNOMED CT. (a)~The Euclidean embedding space of an encoder-based pre-trained language model; semantically relevant concepts are clustered together.  (b)~The Poincaré ball embedding space of the encoder-based language model re-trained by the taxonomy using HiT; Broad concepts (e.g.,\textit{Clinical finding}) are situated closer to the origin and more specific concepts (e.g.,  \textit{Pins and needles}) are positioned toward the boundary.
  (c)~Demonstration the leraning procedure of HiT, where concepts of a subsumption are pulled together while siblings and unrelated concepts are pushed apart, and more general concepts are pulled closer to the Poincaré ball origin.
  }
  \label{fig:hit-embedding-intuition}
\end{figure}

The hyperbolic clustering loss, defined below with hyperbolic margin $\alpha$, acts to pull related concepts together, while pushing unrelated concepts (sampled negatives) apart:
\begin{equation}
\mathcal{L}_{cluster} = \sum_{(x,\ x^{+},x^{-})} \mathrm{max}\Big(0,\ d_\kappa(\boldsymbol{x},\ \boldsymbol{x}^+) - d_\kappa(\boldsymbol{x}^{ }, \boldsymbol{x}^-) + \alpha\Big).
\label{eq:hyperbolic-clustering-loss}
\end{equation}
The hyperbolic centripetal loss, defined below with hyperbolic norm margin $\beta$), situates high-level concepts nearer to the origin:
\begin{equation}
\mathcal{L}_{centri} = \sum_{(x,\ x^{+},x^{-})} \mathrm{max}\Big(0,\ \Vert \boldsymbol{x}^+ \Vert_\kappa - \Vert \boldsymbol{x} \Vert_\kappa + \beta\Big).
\label{eq:hyperbolic-centripetal-loss}
\end{equation}
The total loss, $\mathcal{L}_{\textrm{HiT}}$ is the linear combination of the pair, given by:
\begin{equation}
\mathcal{L}_{HiT} = \mathcal{L}_{cluster} + \mathcal{L}_{centri}.
\label{eq:hit-loss}
\end{equation}
The insight of learning in HiT is demonstrated in Figure~\ref{fig:hit-embedding-intuition-c}.

In addition to modelling taxonomic hierarchies, OnT \cite{yang2025language} extends HiT by incorporating role embeddings and complex $\mathcal{EL}$-concepts. 
Specifically, OnT encodes a role $r$ by a rotation (denoted as $f_r(\cdot)$), and embed a complex concept $D$ by its verbalisation (denoted $\boldsymbol{x}_D$).\footnote{The verbalisation of a complex concept is to transform it into a natural language description with the same semantics using a pre-defined template \cite{he2023language}. For example, to verbalise an existential restriction $\exists r. D$, we use the template ``\textit{something that $\mathcal{V}(r)$ some $\mathcal{V}(D)$}'' where $\mathcal{V}()$ denotes the label of an entity. See the tutorial \url{https://krr-oxford.github.io/DeepOnto/verbaliser/} for more details.} 
Besides the hierarchy loss of the given axioms following HiT, OnT also introduces two different losses for the two logical operators, existential restriction ($\exists r.$) and conjunction ($\sqcap$), used to construct complex $\mathcal{EL}$ concepts, as detailed below.

OnT provides two different embeddings for an existential restriction $\exists r. D$:  $\boldsymbol{x}_{\exists r.D}$ by directly feeding the verbelization of $\exists r. D$ into the language model, or $f_r(\boldsymbol{x}_D)$ by applying the role specific rotation function over the language model embedding of $D$. 
OnT aligns the two embeddings by interpreting their equivalence as two partial-order relationships, and thus the implemented role loss of OnT is decomposed as two bidirectional hierarchy losses as shown below:
\begin{equation}
\mathcal{L}_{r}(\exists r.D) 
= 
\frac{1}{2} 
  \Big(  
    \mathcal{L}_{HiT} 
    \big( 
      \boldsymbol{x}_{\exists r.D}, 
      f_r(\boldsymbol{x}_D)
    \big) 
    + 
    \mathcal{L}_{HiT} 
    \big( 
      f_r(\boldsymbol{x}_D),
      \boldsymbol{x}_{\exists r.D}
    \big) 
  \Big)
.
\label{eq:exist-restrict-loss}
\end{equation}

For conjunctive axioms of the form $C \sqcap D$, the complex concept is necessarily subsumed by each conjunct, i.e., $C \sqcap D \sqsubseteq C$ and $C \sqcap D \sqsubseteq D$. Thus the conjuctive loss is defined as:

\begin{equation}
\mathcal{L}_{\sqcap}(C \sqcap D) 
= 
\frac{1}{2} 
\Big(  
  \mathcal{L}_{HiT}
  \big(
    \boldsymbol{x}_{C \sqcap D}, 
    \boldsymbol{x}_C
  \big) 
  + 
  \mathcal{L}_{HiT}
  \big(
    \boldsymbol{x}_{C \sqcap D}, 
    \boldsymbol{x}_D
  \big) 
\Big)
.
\label{eq:conjunction-loss}
\end{equation}

\textbf{Subsumption Inference with the Embeddings}. After training, both HiT and OnT can be used for subsumption inference using hyperbolic distance and depth-biased scoring, denoted $d_\kappa$ and $s(C \sqsubseteq D)$, which we repurpose for hierarchical retrieval. The scoring function:

\begin{equation}\label{eq:subs-score}
\mathrm{s}(C \sqsubseteq D) := - \big(d_\kappa(\boldsymbol{x}_C, \boldsymbol{x}_D) + \lambda (\|\boldsymbol{x}_{D}\|_{\kappa} - \| \boldsymbol{x}_{C}\|_{\kappa})\big),
\end{equation}

\noindent estimates subsumption confidence, where $\boldsymbol{x}_C$ and $\boldsymbol{x}_{D}$ represent the embeddings of the prospective child and parent $C$ and $D$, respectively; $\lambda$ is a weight determined by the best performance on validation sets and $\|\cdot\|_{\kappa}$ denotes the hyperbolic norm with curvature $\kappa$.

\section{Methodology}
\label{sec:task-definition-and-methodology}

\subsection{Problem Definition}
\label{sec:problem-definition}

In this study, we consider SNOMED CT queries, which can be regarded as keywords or phrases that indicate entity mentions. We begin by defining Hierarchical Retrieval (HR) with such queries over a general ontology.  
\begin{definition} [Ontology Hierarchical Retrieval (HR)] 
Let $\mathcal{O}$ be an ontology consisting of a set of named concepts  $\mathcal{C}$. Given a query $q$ indicating a concept $C_q$ which may or may not belong to $\mathcal{C}$, the task of \textbf{hierarchical retrieval} is to identify the set of concepts $C_1, C_2, \ldots, C_n \in \mathcal{C}$ that subsume $C_q$.
%
The query $q$ is defined as \textbf{out-of-vocabulary} (OOV) if its indicated concept $C_q$ does not exist in $\mathcal{C}$, and this problem becomes HR with an OOV query.
\end{definition}
Note that $C_q$ is expected to be an answering concept whose meaning is exactly indicated by the query $q$. Namely, the semantics $C_q$ is expected to be equivalent to that of $q$. We call $C_q$ as \textit{query indicated answer or concept}. In most ontology retrieval scenarios, such answers do not exist as the concept categorisation may not be fine-grained enough, and the query is often not specifically or clearly expressed. 

In this ontology HR problem, an answer $C_i$ is defined as \textbf{optimal} if it is maximally specific, i.e., there exists no other answer $C_j \in \mathcal{C}$ such that $\mathcal{O}\models C_j \sqsubseteq C_i$ ($j\neq i$). Namely, the optimal answer is the most specific subsumer in the ontology that subsumes the query indicated concept. There may be multiple optimal answers for a query.
Based on this, we can further define two settings for this ontology HR problem:
\begin{enumerate}
  \item \textbf{Optimal Target}: Only the optimal answers are expected to be returned. They are denoted as $Ans^\star(q)$.
  \item \textbf{Multi-Hop Target}: Both optimal answers $Ans^\star(q)$ and their ancestors within $d$ hops in the concept hierarchies of the ontology are expected to be returned as illustrated in Figure \ref{fig:hierarchy-fragment}. These answers are denoted as $Ans_{\leq d}(q)$.   
  Specifically, the $d$-hop neighbours of $C$ are the nodes $D$ whose shortest path distance from $C$ in the concept hierarchy is not larger than $d$. Formally, this means that there exists a shortest inference chain
\[
C = C_0, C_1, \ldots, C_d = D
\]
such that $\mathcal{O} \models C_i \sqsubseteq C_{i+1}$ for each $i$, where each subsumption is \emph{direct}---that is, there is no intermediate concept $C'$ such that
\[
\mathcal{O} \models C_i \sqsubseteq C' \quad \text{and} \quad \mathcal{O} \models C' \sqsubseteq C_{i+1}.
\] 
\end{enumerate}
Specially, $Ans_{\leq 0}(q) = Ans^\star(q)$. 

\subsection{Our Approach} 
\label{sec:approach}

\begin{figure}[htb!]
  \centering
  \includegraphics[width=\columnwidth]{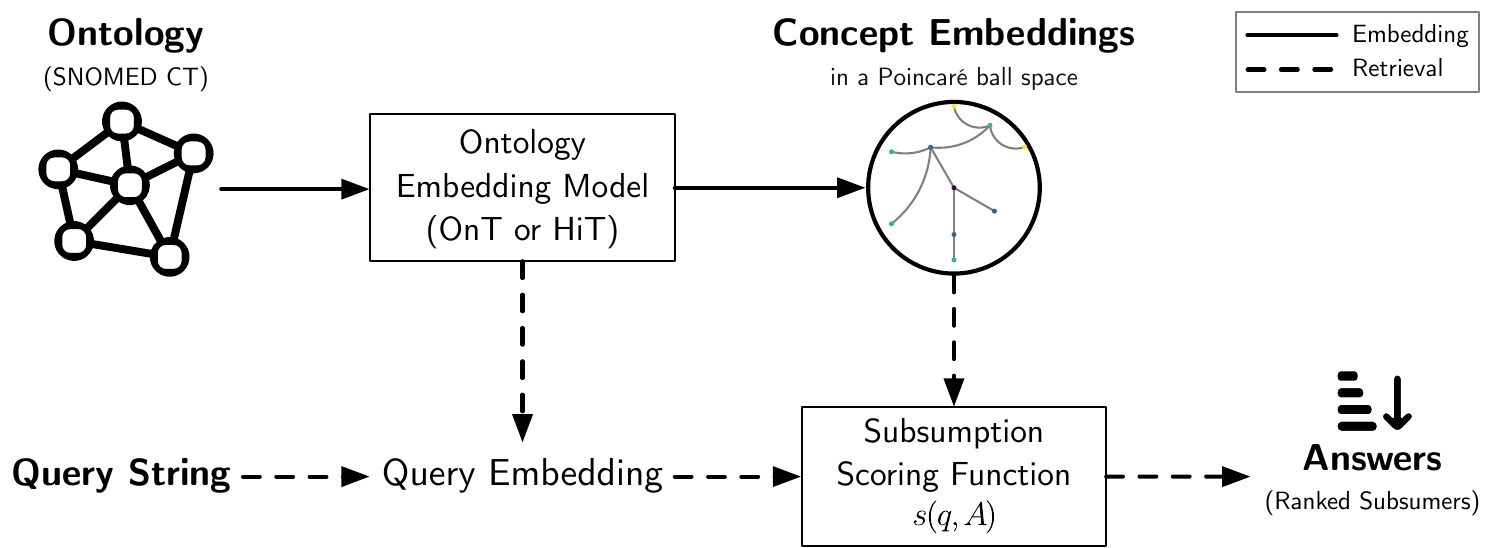}
  \caption{Architecture of our ontology embedding-based approach for SNOMED CT hierarchical retrieval. Both ontology embedding models, HiT and OnT, can be applied. 
  }
  \label{fig:approach-overview}
\end{figure}

Our HR approach for SNOMED CT with OOV queries is composed of two phases. First, all SNOMED CT concepts are embedded (solid arrows in Fig.~\ref{fig:approach-overview}) using their textual class labels and a trained ontology embedding model, creating an embedding store. Then, during retrieval (dashed arrows in Fig.~\ref{fig:approach-overview}), the query string is encoded by the same ontology embedding model and scored against all pre-computed concept embeddings, yielding a list of concept candidates ranked according to their likelihood of being the query's subsumer, and the top-k ranked concepts are returned as the retrieval results.

For computing SNOMED CT concept embeddings, we re-train a language model as an encoder using either HiT or OnT, which produces the concept embeddings in a hyperbolic space.
In retrieval, we treat all the named concepts in SNOMED CT as candidate concepts.
For our approach with different embedding models, and the baselines, we use the same concept labels. {For each concept, we use its English label annotated by \texttt{rdfs:label}, and preprocess it with the following steps: removing SNOMED CT specific branch tag which is attached to the concept name with brackets for indicating which branch the concept belongs to\footnote{One example of such branch tag is ``(finding)'' in the concept label ``Finding by site (finding)''; the tag is not a part of the concept name, but used for visualisation and easier human access.}, removing punctuation, and converting all the letters to lowercase (detailed in Section~\ref{sec:dataset-construction-eval-100}).
}
The input query $q$ is embedded (denoted as $\boldsymbol{x}_q$) using the same trained encoder for ontology embedding, and scored against pre-computed concept embeddings using the standard subsumption inference scoring function of HiT and OnT (i.e., Equation \eqref{eq:subs-score}).
Namely, for an named concept $A$ whose embedding is denoted as $\boldsymbol{x}_A$, its score of being one answer of $q$ (i.e., a subsumer of the query indicated concept) is calculated as:
\begin{equation}
\mathrm{s}(q, A) := - \big(d_\kappa(\boldsymbol{x}_q, \boldsymbol{x}_A) + \lambda (\|\boldsymbol{x}_{q}\|_{\kappa} - \| \boldsymbol{x}_{A}\|_{\kappa})\big),
\label{eq:problem-specific-subs-score}
\end{equation}

\noindent where $\lambda$ is selected based on the validation set as detailed in Section~\ref{sec:hyperparameter-selection}.

\section{Datasets}
\label{sec:datasets}

In this section, we introduce the three datasets Eval-100, OET-CCP and OET-Disease that are used for evaluation, along with their construction details.

\subsection{Eval-100}
\label{sec:dataset-construction-eval-100}

Eval-100 includes 100 OOV queries drawn from the MIRAGE benchmark\footnote{\url{https://github.com/Teddy-XiongGZ/MIRAGE/blob/main/benchmark.json}} \cite{xiong-etal-2024-benchmarking}, 
which consists of $7663$ biomedical questions written in both layman and clinically precise styles. 
We manually create the queries and annotate their ground truth answers with named concepts in the international version of SNOMED CT released in September 2025. The construction of Eval-100, which includes two continuous stages -- generation of candidate OOV queries and manual assessment of canddiate OOV queries, is described below.

\begin{figure}[htb!]
  \centering
  \includegraphics[width=\columnwidth]{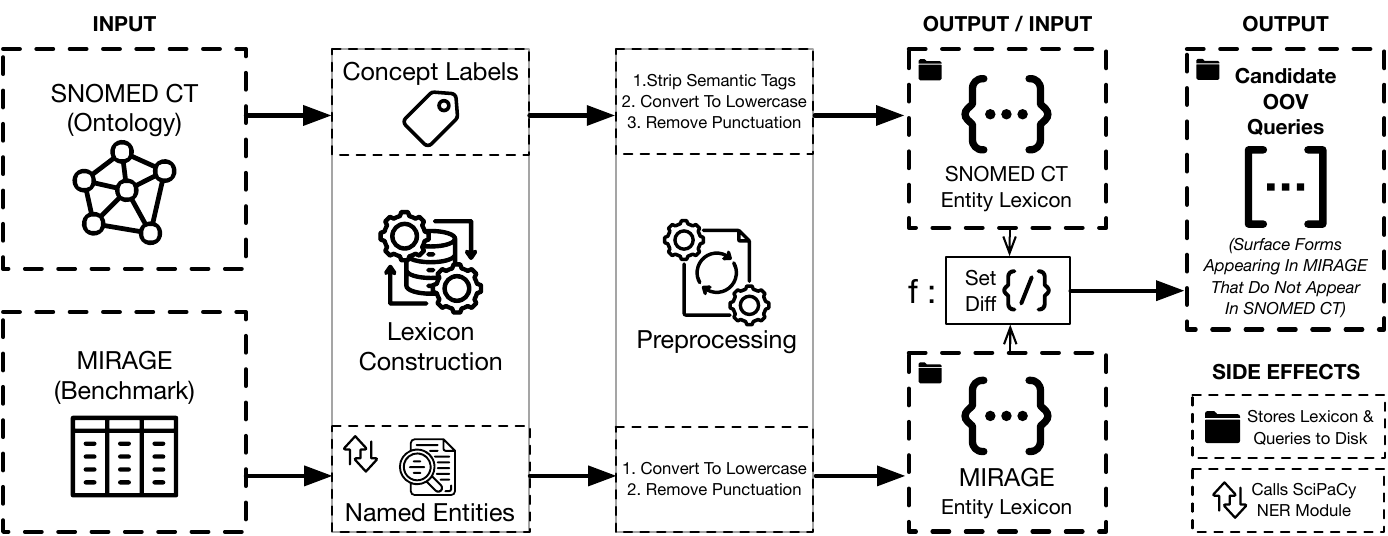}
  \caption{Candidate OOV query generation which is mainly based on entity extraction from MIRAGE questions, and entity comparison with SNOMED CT concepts.
  }
  \label{fig:lexicon-construction-with-oov-query-candidates}
\end{figure}

\paragraph*{Generation of Candidate OOV Queries} \label{sec:lexicon-construction-mirage-and-snomed} 
As shown in Fig.~\ref{fig:lexicon-construction-with-oov-query-candidates}, we generate query candidates by (1) extracting entities from questions in the MIRAGE benchmark, and (2) comparing these entities with concepts in SNOMED CT, where those entities that differ lexically from any of the concepts in SNOMED CT are selected as OOV query candidates. 
To extract entities from MIRAGE questions, we use SciSpaCy\footnote{The \texttt{en\_ner\_bionlp13cg\_md} model, found at \url{https://allenai.github.io/scispacy/}, is used.} to analyse the question text and conduct named entity recognition. SciSpaCy is selected as it is good at processing biomedical text and also provides each mention with a categorisation label, which can help with annotation in the next manual assessment step. Each extracted entity mention is stripped of punctuation and converted to lowercase, resulting in our MIRAGE entity lexicon.
For SNOMED CT, we extract and process the English labels annotated by \texttt{rdfs:label} for all its named concepts. The processing includes removing branch tags and punctuation, and converting to lowercase.
This provides our SNOMED~CT entity lexicon. 
With the MIRAGE entity lexicon and SNOMED CT entity lexicon, we take a set difference operation between them, and get all the entity mentions from MIRAGE whose surface form does not appear in  any SNOMED~CT concept label. This yields 3,530 candidate OOV queries.

\begin{example} \label{ex:eval-100-0}
The entity ``cold induced tingling in fingers'' is extracted from a question ``A 62-year-old woman comes to the physician because of a 6-month history of progressive pain and stiffness in her right hand with a cold induced tingling in fingers. Which of the following is the most likely cause of her underlying symptoms?''. It is not matched with any label of the SNOMED CT concepts and is selected as a candidate OOV query. 
\end{example}

\paragraph*{Manual Assessement of Candidate OOV Queries} \label{sec:annotation-procedure}

\begin{figure}[htb!]
  \centering
  \includegraphics[width=0.9\columnwidth]{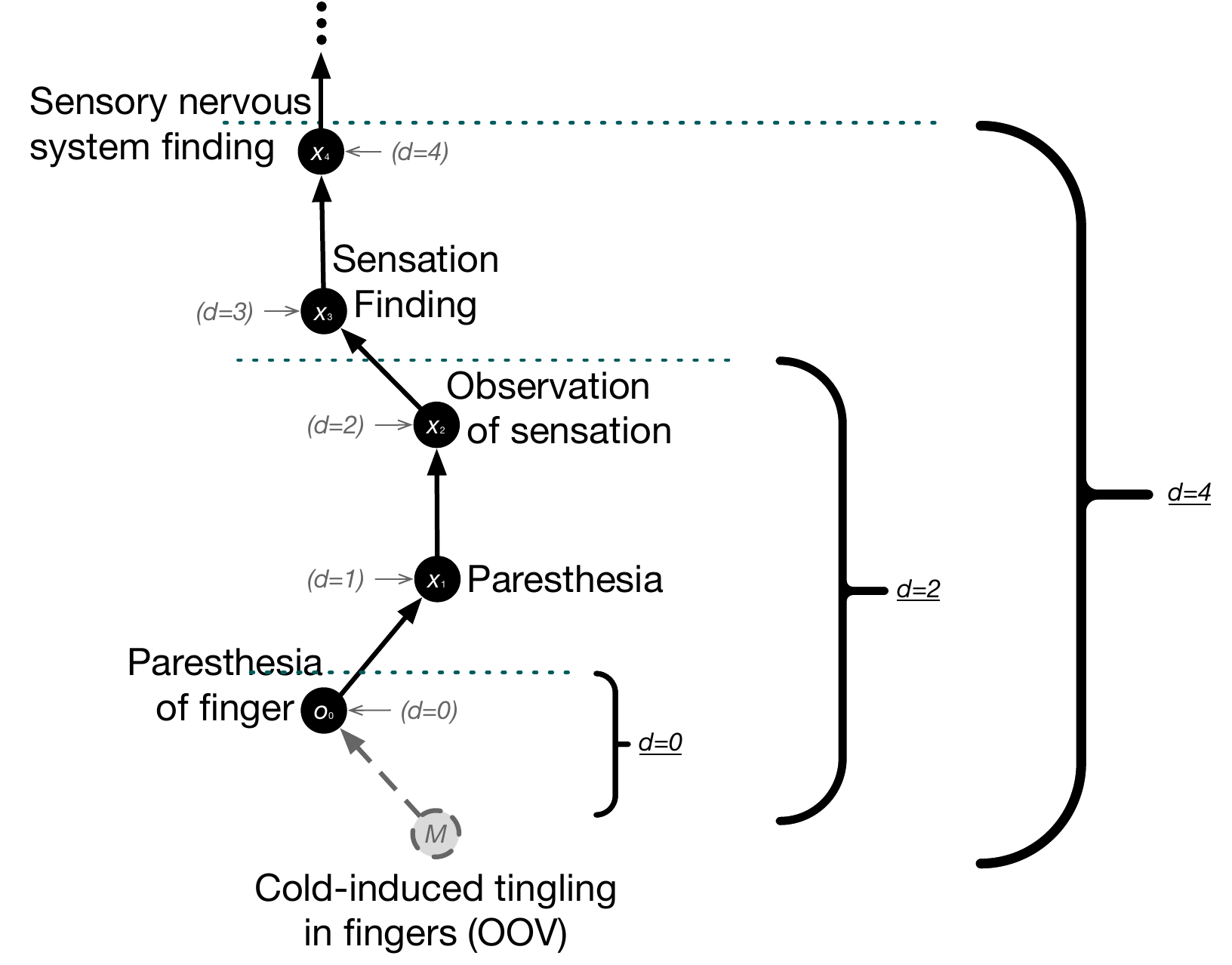}
  \caption{Demonstration of the optimal and multi-hop targets of an example query.
  }
  \label{fig:oov-query-target-ancestral-chain}
\end{figure}

For each candidate query, the human annotator manually assesses its qualification and annotates its targets (answers) with the following four steps.
\begin{enumerate}
    \item \textbf{Synonym-based filtering:} The annotator checks and ensures that the query is lexically different from any concept synonym annotated by \texttt{altLabel} in SNOMED~CT.  Otherwise, this candidate query is discarded.
    \item \textbf{Semantic equivalence-based filtering:} The annotator accesses concepts that are relevant to the query in SNOMED CT based on string matching\footnote{The concepts whose labels have a high similarity w.r.t. the query based on Jaccard similarity and word-level containment are regarded as relevant.}, and manually assesses whether each relevant concept is semantically equivalent to the query based on its context in the hierarchy. If an equivalent concept is found, the query is discarded. 
    \item \textbf{Optimal target annotation:} If a query is kept in the above two steps, the annotator annotates its optimal targets (answers). As in the above step, the annotator finds relevant concepts and searches their contexts in SNOMED CT with the support of the tool Protégé for the most specific subsumers. If reasonable most specific subsumers are found, this query is kept together with the found optimal answers. Otherwise, the query is discarded. 
   \item \textbf{Multi-hop target annotation:} All the ancestors of each optimal target are inferred in SNOMED CT. Except for \textit{owl:Thing}, these ancestors are kept as multi-hop targets, and their depths to the corresponding optimal target in the concept hierarchy are recorded. 
\end{enumerate}
The annotation process is repeated for each candidate query, and stops when 100 qualified OOV queries have been successfully annotated. These queries, as well as their optimal and multi-hop targets, compose the Eval-100 benchmark.
\begin{example} \label{ex:eval-100-1}
The candidate query ``cold induced tingling in fingers'' in Example \ref{ex:eval-100-0} is kept in the first two assessment steps, and in the third step, it is annotated with the most specific subsumer of  \textit{\textit{Paresthesia of finger (finding)}}. As \textit{Paresthesia of finger (finding)} is directly subsumed by \textit{Paresthesia (finding)}, \textit{Paresthesia (finding)} is kept as a mult-hop target within $Ans_{\leq d}(q)$ ($d\ge1$). The example query, its optimal target and more multi-hop targets can be found in Fig~\ref{fig:oov-query-target-ancestral-chain}.
\end{example}

\subsection{OET-CPP and OET-Disease}
\label{sec:dataset-construction-OET}

In this section, we introduce two new OOV query datasets, \textbf{OET-CPP} and \textbf{OET-Disease}, derived from the existing \textit{Ontology Enrichment from Text (OET)} dataset~\cite{dong2023ontology}. 
The OET dataset is designed for the task of taxonomy insertion, which aims to place a new concept into an existing taxonomy by identifying its appropriate parent and child concepts. 
For obtaining those new concepts,  OET compares different versions of SNOMED CT (2017 and 2014) and treats terms that appear only in the newer version as the new concept. 

We construct our OOV query dataset by adopting the OET dataset as follows: we treat the new concept as the OOV query and consider its correct parent in the taxonomy as the optimal target for the query. 
Then, we extend the optimal target case to Multi-hop cases by computing the chain of ancestor concepts  as in section~\ref{sec:annotation-procedure}. For simplicity and to be consistent with the settings in Section~\ref{sec:annotation-procedure}, we ignore the child annotations, and omit  extra datas provided  the OET data, such as left and right context.

It is worth noting that the SNOMED CT versions play an important role in the construction of the OET dataset, and could also have important implications for model training. Specifically, for preventing data leakage, we must train version-specific encoders to evaluate retrieval performance on this adapted dataset, as further discussed in section~\ref{sec:model-variants}.

\paragraph*{Data Extraction and Preprocessing}

To remove redundant or irrelevant data, we perform the following preprocessing steps. 
First, we remove short queries (i.e., queries with a number of characters less than or equal to 5) since these are either acronyms or previously relied on the included context for disambiguation during placement. 
Also, for avoiding queries with a general and infoless answer (e.g., $Disease$), we remove the query with answers having depth less than $5$ in the SNOMED CT Taxonomy.

Finally, we procedure two distinct datasets by applying the procedure above to the two partitions of the original OET datasets: (1) \textbf{OET-CPP}, containing target concepts from the semantic branches of \textit{Clinical finding}, \textit{Pharmaceutical}, and \textit{Procedure}; and (2) \textbf{OET-Disease}, containing target concepts only from the disease branch of SNOMED CT. 
By abuse of notations, we still call it  OET-CPP and  OET-Disease \footnote{Available at \url{https://github.com/jonathondilworth/HR-OOV-SNOMED-CT/tree/main/data}.}.

\subsection{Dataset Statistics}
\label{sec:dataset-statistics}

\begin{table}[htbp]
    \begin{center}
        \caption{Statistics of the evaluation dataset for the optimal target case (i.e., $d=0$) and the multi-hop target case (i.e., $d>0$). The multi-hop target setting includes $d \in \{0,2,4\}$ as well as the unbounded case $d=\infty$. The \textit{Avg. Depth} represents the average height of each query's ontology fragment (i.e., the hop-based distance from $q$ to the most distant ancestor).}
        \label{tab:dataset-stats}
        \begin{tabular}{@{}lcccccccc@{}}
            \toprule
            & & & & \multicolumn{4}{c}{Mean Target Count} & Avg. \\
            \cmidrule(lr){5-8}
            Dataset & Pairs & Multi & Queries & $d=0$ & $d=2$ &  $d=4$ & $d=\infty$ & Depth \\
            \midrule
            Eval-100\textsuperscript{\textasteriskcentered} & 100 & 0 & 100 & 1.0 & 4.98 & 10.78 & 17.40 & 7.90 \\
            OET-CPP\textsuperscript{\textdagger} & 202 & 26 (15.6\%) & 167 & 1.21 & 3.86 & 6.05 & 7.89 & 6.89 \\
            OET-Disease\textsuperscript{\textdagger,\textdaggerdbl} & 80 & 16 (29.1\%) & 55 & 1.45 & 4.51 & 6.98 & 9.69 & 8.51 \\
            \bottomrule
        \end{tabular}
    \end{center}
    {\small
    \textsuperscript{\textasteriskcentered}Manually annotated, SNOMED CT (version Sep.\ 2025).
    \textsuperscript{\textdagger}KB-versioning, SNOMED CT (version Sep.\ 2017 $-$ Sep.\ 2014).
    \textsuperscript{\textdaggerdbl}Domain-restricted subset of OET-CPP; 78 of 80 pairs overlap.
    }
\end{table}

Recall that Eval-100 queries correspond to named entities drawn from biomedical questions within the MIRAGE benchmark, and OET-variant queries originate from PubMed abstracts and are organised via the OET concept placement datasets, i.e., CPP and Disease. It is worth noting that, we exclude OET-variant queries with optimal targets within 5 hops of \texttt{owl:Thing}, to avoid the trivial and meaningless case with \texttt{owl:Thing} as a candidate in the multi-hop target case. 
For simplicity, we also ignore the datas in the original OET containing complex concepts, whose
 optimal targets are typically shallow in the ontology, so removing them 
has little impact on the overall data quality. After the two removals, the size of the final resulting  datasets OET-CPP (originally 2131 out-of-KB mention-edge pairs) and OET-Disease (originally 1637) are reduced to 202 and 80, respectively.

All evaluation queries across the three datasets are roughly 2--2.5 words in length (the mean query word count for EVAL-100, OET-CPP, and OET-Disease is 2.10, 2.23, and 2.45, respectively). OET-Disease exhibits a higher proportion of queries with multiple optimal targets (i.e., OOV queries that represent prospective concepts with $> 1$ parent) at 29.1\% compared with OET-CPP at 15.6\%. We include the unbounded target count, i.e.\ $d=\infty$, to contextualise the bounded cases for $d \in \{1,3,5\}$. For instance, the typical query from Eval-100 covers only 62\% of its entire ancestral chain (10.84 of 17.47) at $d=4$, whereas this value is 77\% (6.05 of 7.89) for OET-CPP and 72\% (6.98 of 9.69) for OET-Disease. 

Although the OET mentions and their optimal targets are originally obtained from OET's pruned ontologies, we compute the ancestral chains for OET-CPP and OET-Disease against the inferred subsumption relation of the September 2014 US release, reusing the same pipeline as in Eval-100. The difference in target surface (i.e., the mean target count at $d=\infty$) is attributable to structural and completeness variations across the ontology versions. The SNOMED CT international release from September 2025 is substantially larger than the US release from September 2014, containing more intermediate concepts and relational properties. This increases the number of target concepts in each query's ancestral chain for Eval-100 compared to OET-CPP and OET-Disease. For instance, Eval-100 and OET-Disease have similar mean max depths (7.88 and 8.51); however, Eval-100 has nearly twice as many total ancestors (17.47 vs 9.69). These structural differences should be considered when comparing results across datasets in Section~\ref{sec:experimental-results}.

\section{Experiments}
\label{sec:experiments}

\subsection{Experimental Setup}
\label{sec:experimental-setup}

The evaluation is done as follows: 
First, we test the retrieval of the optimal target (i.e., $d=0$) case. This allows us to measure each method's effectiveness for fine-grained hierarchical retrieval. Then, we test the multi-hop target case  with $d \in \{2,4\}$, 
which allows us to assess each method's ability to find correct but less fine-grained answers.   

\paragraph*{Evaluation Metrics} \label{sec:evaluation-metrics} In both the optimal and multi-hop settings, we evaluate the performance using the ranking-based metrics, including mean reciprocal rank (MRR), hit rate (H@k, $k\in\{0,2,4\}$), and mean rank (MR). It is worth noting that when a query has multiple targets, the ranking-based metrics and hit rate are computed using the best rank among all its answers. 

\paragraph*{Baselines}
\label{sec:baselines}

\subparagraph*{Lexical Baselines} \label{sec:lexical-baselines} For lexical baselines, we employ TF--IDF and BM25. For both methods, we build an index over the preprocessed class labels for all SNOMED CT concepts (i.e., each label then constitutes a document). No additional label processing is performed beyond the existing preprocessing steps outlined in section~\ref{sec:datasets}.
For BM25, we tune the hyperparameters $k_1=1.5$ and $b=0.7$ on a set of 30 non-overlapping queries (i.e., disjoint from the evaluation set). By comparing with these lexical baselines, we show the improvements of our methods that go beyond the surface-level lexical overlap.

\subparagraph*{Embedding-based Baselines} \label{sec:embedding-based-baselines} We include the pre-trained language model SBERT \textit{(the version with all-MiniLM-L12-v2)} as one embedding-based baseline. 
It is worth noting that the HiT and OnT models used in our HR methods are fine-tuned based on this SBERT module.
Alongside the base model, we include the domain-tuned encoder, SapBERT \cite{liu2021self}, trained on UMLS synonym pairs using a self-alignment pretraining objective. By comparing our methods against SapBERT tests, we show that whether domain-specific pre-training alone--without hyperbolic and logical objectives--remains competitive with our hierarchical retrieval methods. Both baselines provide contextualised embeddings, scored using cosine similarity.

\begin{table*}[htb!]
    \caption{An overview of each experimental method. All HiT and OnT models use \textit{all-MiniLM-L12-v2} as their base encoder and are hierarchy retrained over 4 epochs. Lexical baselines use whitespace tokenisation over pre-processed concept labels. The subsumption score $s(q, A)$ is used for ranking in the main results; hyperbolic distance results are reported in Appendix.~\ref{appendix:full-experimental-results}.}
    \label{tab:experimental-settings}
    \centering
    \begin{tabular}{llll}
        \toprule
        \textbf{Method} & \textbf{Type} & \textbf{Ranking} & \textbf{Hyperparameters} \\
        \midrule
        \multicolumn{4}{@{}l}{\textit{Baseline Methods}} \\
        TF-IDF  & Lexical   & TF-IDF        & ---                       \\
        BM25    & Lexical   & BM25          & $k_1 = 1.5$, $b = 0.7$   \\
        SBERT   & PLM       & Cosine sim.   & ---                       \\
        SapBERT & PLM        & Cosine sim.   & ---                       \\
        \multicolumn{4}{@{}l}{\textit{Our HR Methods}} \\
        HiT     & Taxonomy  & $s(q, A)$     & $\lambda$ (see Table.~\ref{tab:exp-lambda-settings}) \\
        OnT     & $\mathcal{EL}$-concepts & $s(q, A)$& $\lambda$ (see Table.~\ref{tab:exp-lambda-settings}) \\
        \bottomrule
    \end{tabular}
\end{table*}

\paragraph*{Variants of Our Models} \label{sec:model-variants}

\subparagraph*{Version-specific Encoders:} \label{sec:version-specific-encoders} For the EVAL-100 evaluation, we used the existing HiT and OnT models that are trained on the September 2025 release of SNOMED CT, ensuring consistency between the specific ontology used for training and the concept hierarchy used to obtain each query's ancestor set. However, since the OET-CPP and OET-Disease datasets are derived from the versioning procedure discussed in Section~\ref{sec:dataset-construction-OET}, for a fair evaluation, we retrain HiT and OnT  on the September 2014 release. As the existing HiT/OnT was trained on the SNOMED-25 version, it may identify concepts that do not exist in the 2014 release, which would result in leakage relative to the OET evaluation setting.

\subparagraph*{Variant with different training sets} \label{sec:training-set-variants} To better understand the effects of training set size on the performance of HiT and OnT, we investigate two model variants that are trained on the full ontologies or just a part of them.  For EVAL-100, where we use the 2025 version of SNOMED CT, the part is obtained by selecting several semantic branches: \textit{Body Structure}, \textit{Clinical Finding}, \textit{Event} and \textit{Procedure}. In the case of OET-CPP and OET-Disease, we use the same part as the pruned  SNOMED CT (version  2014) provided by the original OET dataset. 
The details are summerised in Table.~\ref{tab:exp-lambda-settings}, where we represent the two versions that train on full ontologies or part of them as  F and M, respectively.

\paragraph*{Experiment Parameters \& Hyperparameter Selection} \label{sec:hyperparameter-selection} Following HiT \cite{he_language_2024} and OnT \cite{yang2025language}, we train each model using their standard configuration (provided in Appendix.~\ref{appendix:full-experimental-settings}). However, instead of training HiT over 20 epochs and OnT over 1 epoch, we opt to train each model for 4 epochs, as we found this appropriately minimises the loss across both models. For the SNOMED-14 encoders, the value for $\lambda$ is set by hyperparameter tuning on the validation set produced during training, in line with \cite{he_language_2024,yang2025language}. In the case of HiT and Ont in our model, $\lambda$ is selected by tuning over a separate validation set of 30 OOV query-target pairs that optimise $\lambda$ for selecting answers within a depth threshold of as $5$. We provide a description of our experiment settings in Table.~\ref{tab:exp-lambda-settings}.

\begin{table*}[htb!]
    \caption{Encoder configurations and $\lambda$ values for each dataset--model combination. M and F denote miniature and full encoder variants, respectively. For the SNOMED-14 encoders, $\lambda$ is set via hyperparameter tuning on the validation set produced during training~\cite{he_language_2024,yang2025language}. For the SNOMED-25 encoders, $\lambda$ is tuned over a separate validation set of 30 OOV query-target pairs.}
    \label{tab:exp-lambda-settings}
    \centering
    \begin{tabular}{lllc}
        \toprule
        \textbf{SNOMED CT Release} & \textbf{Model} & \textbf{Training Ontology\textsuperscript{\dag}} & $\boldsymbol{\lambda}$ \\
        \midrule
        \multirow{4}{*}{Sep.\ 2014 (v.20140901)}
            & HiT (M) & OET-pruned      & 0.7 \\
            & HiT (F) & Full US release  & 0.8 \\
            & OnT (M) & OET-pruned       & 0.4 \\
            & OnT (F) & Full US release   & 0.5 \\
        \midrule
        \multirow{4}{*}{Sep.\ 2025 (v.20250901)}
            & HiT (M) & Semantic branches\textsuperscript{\ddag} & 0.8 \\
            & HiT (F) & Full int.\ release & 0.4 \\
            & OnT (M) & Semantic branches\textsuperscript{\ddag} & 0.6 \\
            & OnT (F) & Full int.\ release  & 0.6 \\
        \bottomrule
    \end{tabular}

    \smallskip
    \noindent\footnotesize\textsuperscript{\dag}During training $\alpha=3.0,\ \beta=0.5$.\textsuperscript{\ddag}Body Structure, Clinical Finding, Event, and Procedure.
\end{table*}

\subsection{Experimental Results}
\label{sec:experimental-results}

The overall performance of different methods on the HR task over all three datasets is summarised in Table~\ref{tab:results-all}. 
We observe that our methods based on hierarchy embeddings achieve overall better performance than baseline methods that focus on similarity computations, particularly in the multi-hop target setting. 
For instance, the best-performing HiT model at $d=4$ achieves MRR scores of 54 and 50 on OET-CPP and OET-Disease, respectively, substantially outperforming the best SapBERT baseline (54 vs. 26 and 50 vs. 39); 
This pattern remains consistent for HiT and OnT across multi-hop cases (i.e., $d \in \{2,4\}$), which is reasonable as HiT and OnT are trained in hyperbolic space and designed with a better capability for capturing hierarchical structures. 

\paragraph*{Optimal Target Setting}
In the optimal target setting (i.e., $d=0$), our methods, HiT and OnT, show overall the best and most stable performance among all baseline methods. The only exception occurs on the OET-Disease dataset, where SapBERT achieves better results in terms of MRR and MR (38 vs. 30 for MRR and 1300 vs. 2561 for MR).

However, on the EVAL-100 and OET-CPP datasets, SapBERT exhibits much worse MR performance than HiT and OnT, with MR values approximately 6–10 times larger (6667 vs. 1170 and 26942 vs. 2179). This indicates that our methods are more stable and better able to handle difficult cases that may lead to very large ranks and consequently higher mean rank values.

In terms of other metrics, HiT, OnT, and SapBERT achieve similar performance. The other methods, especially the lexical-based approaches, perform substantially worse, with MR values even exceeding 250,000. Moreover, the original SBERT model without fine-tuning performs better than lexical-based methods but still underperforms the fine-tuned models by about 10 points in terms of H@K metrics.

\begin{table*}[htb!]
  \caption{Retrieval performance across all three evaluation datasets and depth threshold. HiT and OnT results use SNOMED-25 and SNOMED-14 encoders for EVAL-100 and OET variant datasets, respectively. HiT and OnT encoders use the subsumption score $s(q,A)$ for ranking. A full set of measures (with hyperbolic distance included) are reported in Appendix.~\ref{appendix:full-experimental-results}. M and F under method denote miniature and full encoder variants, respectively. MRR and H@k (with $k=\{1,3,5\}$) are to be read as percentages. \underline{\textbf{Underlined bold text}} signifies the best performance. \textbf{Bold text only} signals a draw.}
  \label{tab:results-all}
  \begin{center}
    \small
    \setlength{\tabcolsep}{4.5pt}
    \begin{tabular}{@{} l ccc ccc ccc @{}}
      \toprule
      & \multicolumn{3}{c}{\textbf{EVAL-100}} 
      & \multicolumn{3}{c}{\textbf{OET-CPP}} 
      & \multicolumn{3}{c}{\textbf{OET-Disease}} \\
      \cmidrule(lr){2-4} \cmidrule(lr){5-7} \cmidrule(lr){8-10}
      Method 
        & MRR & H@k & MR 
        & MRR & H@k & MR 
        & MRR & H@k & MR  \\
      \midrule
      \multicolumn{10}{@{}l}{\textit{Optimal Target ($d=0$)}} \\
      \midrule
      TF-IDF   & 04 & 02/03/04 & 251820  & 09 & 07/10/11 & 107927  & 12 & 07/15/16 & 82027 \\
      BM25     & 06 & 05/06/07 & 101885  & 10 & 08/11/11 & 51081   & 12 & 09/11/13 & 30056 \\
      SBERT    & 11 & 05/14/16 & 20127   & 17 & 10/20/24 & 12683   & 23 & 16/29/31 & 3804  \\
      SapBERT  & \textbf{14} & 09/14/\underline{\textbf{20}} & 26942   & \textbf{25} & \textbf{17}/28/34 & 6667    & \underline{\textbf{38}} & \underline{\textbf{27}}/\textbf{40}/49 & \underline{\textbf{1300}}  \\
      \addlinespace
      HiT (M)  & 06 & 05/05/06 & 3773    & 17 & 10/18/22 & 1292    & 20 & 15/24/24 & 3057  \\
      HiT (F)  & \textbf{14} & 09/15/17 & 2647    & 11 & 06/11/14 & 1827    & 11 & 07/13/15 & 3855  \\
      OnT (M)  & \textbf{14} & \underline{\textbf{10}}/15/17 & 2494    & \textbf{25} & 14/\underline{\textbf{29}}/\underline{\textbf{38}} & 1339    & 30 & 16/\textbf{40}/\underline{\textbf{51}} & 3443  \\
      OnT (F)  & 13 & 06/\underline{\textbf{16}}/19 & \underline{\textbf{2179}}    & \textbf{25} & \textbf{17}/28/31 & \underline{\textbf{1170}}    & 31 & 22/38/40 & 2561  \\
      \midrule
      \multicolumn{10}{@{}l}{\textit{Multi-Hop Target ($d=2$)}} \\
      \midrule
      TF-IDF   & 04 & 02/04/05 & 244291  & 10 & 07/12/13 & 105832  & 13 & 07/16/18 & 75686 \\
      BM25     & 07 & 05/07/08 & 75377   & 11 & 09/12/13 & 42459   & 14 & 11/13/15 & 23499 \\
      SBERT    & 12 & 05/15/17 & 11994   & 19 & 11/23/28 & 3923    & 28 & 18/36/38 & 1282  \\
      SapBERT  & 15 & 09/15/20 & 19456   & 26 & 18/29/36 & 3552    & 39 & 27/42/55 & 1089  \\
      \addlinespace
      HiT (M)  & 11 & 08/10/13 & 1163    & 38 & 25/45/52 & 52      & 36 & 24/47/53 & \underline{\textbf{59}}    \\
      HiT (F)  & \underline{\textbf{22}} & \underline{\textbf{15}}/\underline{\textbf{22}}/\underline{\textbf{30}} & 722     & 33 & 19/41/50 & 55      & 20 & 11/22/27 & 88    \\
      OnT (M)  & 18 & 12/19/22 & \underline{\textbf{577}}     & 38 & 24/\underline{\textbf{46}}/\underline{\textbf{54}} & \underline{\textbf{46}}      & \underline{\textbf{44}} & 25/\underline{\textbf{56}}/\underline{\textbf{65}} & 60    \\
      OnT (F)  & 16 & 06/20/26 & 646     & \underline{\textbf{39}} & \underline{\textbf{27}}/43/49 & 51      & 41 & \underline{\textbf{29}}/47/51 & 67    \\
      \midrule
      \multicolumn{10}{@{}l}{\textit{Multi-Hop Target ($d=4$)}} \\
      \midrule
      TF-IDF   & 05 & 02/04/06 & 244290    & 10 & 07/12/13 & 105820  & 13 & 07/16/18 & 75651 \\
      BM25     & 07 & 05/07/08 & 64054     & 12 & 10/13/13 & 41353   & 14 & 11/13/15 & 21960 \\
      SBERT    & 12 & 05/15/17 & 10699     & 19 & 11/23/28 & 3129    & 28 & 18/36/38 & 793   \\
      SapBERT  & 15 & 09/15/20 & 14997     & 26 & 18/29/36 & 2898    & 39 & 27/42/55 & 968   \\
      \addlinespace
      HiT (M)  & 16 & 11/14/20 & 733     & 49 & 33/59/66 & 12      & \underline{\textbf{50}} & 35/\underline{\textbf{62}}/65 & \textbf{11}    \\
      HiT (F)  & \underline{\textbf{24}} & \underline{\textbf{16}}/\underline{\textbf{25}}/\underline{\textbf{33}} & 392     & \underline{\textbf{54}} & \underline{\textbf{41}}/\underline{\textbf{62}}/\underline{\textbf{70}} & \underline{\textbf{11}}      & 46 & \underline{\textbf{36}}/49/56 & 17    \\
      OnT (M)  & 20 & 13/22/26 & \underline{\textbf{303}}     & 42 & 26/49/59 & 14      & 46 & 25/60/\underline{\textbf{71}} & \textbf{11}    \\
      OnT (F)  & 18 & 06/21/29 & 317     & 47 & 32/54/63 & 15      & 47 & 31/53/65 & 23    \\
      \bottomrule
    \end{tabular}
  \end{center}
\end{table*}

\paragraph*{Multi-Hop Target Setting}
On the multi-hop target case, our hierarchical retrieval methods show a substantially better performance than all baselines across  all metrics.
For instance, when $d=4$, the MRR of HiT on OET-CPP doubles compared to the best-performing baseline, SapBERT. Similarly, when $d=4$, the MR of OnT on EVAL-100 is around 30 times better than that of the best-performing SBERT model. 
Moreover, we observe consistent improvements for both HiT and OnT as the hop distance $d$ increases. For example, the MR of HiT on EVAL-100 decreases by approximately five times (3773 vs. 733) when $d$ increases from 0 to 4. A similar trend is observed for OnT on OET-CPP, where the MR decreases by nearly 100 times (1170 vs. 15). 
These results suggest that by leveraging hierarchical embeddings, our methods are highly effective at retrieving related multi-hop targets.

In contrast, all baseline methods have similar answers ranked across every depth threshold, suggesting that they are not capable of distinguishing the answer in different hops. In particular, all lexical and embedding-based baselines plateau in MRR at $d=2$ and as $d$ increases to $4$, only marginal gains in MR and H@k performance are observed. Specifically, there is only a 1 to 2-point gain in MRR across all baselines between $d=0$ and $d=4$, with the exception of SBERT on OET-disease, which achieves a 5-point gain. Meanwhile, the relative gains for HiT and OnT range from a minimum of 5 points, OnT (F) on EVAL-100, to 43 points; i.e., HiT (F) on OET-CPP advances from an MRR of 11, one of the worst-performing variants, to 54, the most performant.

\paragraph*{Training Data Size Influence}
The influence of the training data size does not appear consistent. In many cases, we find that training on a smaller part may even lead to better performance. For instance, on OET-CPP, HiT (M) obtains MRR=17 at $d=0$, outperforming HiT (F), which scores 11, then at $d=4$ this reverses to 49 and 54, respectively. Moreover, the MRR scores of HiT (M) may be slightly better than the full version HiT (F).
This suggests that the standard training procedure of HiT and OnT may suffer from a forgetting problem when handling large datasets, potentially leading to the loss of previously learned information.

\subsection{Discussion}
\label{sec:discussion}

First, it is important to note that unlike EVAL-100, the OET-CPP and OET-Disease datasets were not subjected to manual annotation to enforce lexical and semantic disjointness.  By manually reviewing, we found that 11 samples (5.4\%) in OET-CPP and 7 samples (8.75\%) in OET-Disease could be considered semantically equivalent according to the criteria used for EVAL-100. Moreover, while EVAL-100 shows 0.0\% lexical overlap across all OOV query-target pairs, OET-CPP and OET-Disease show 8.91\% and 5.00\%, respectively. These may contribute to some reason why for $d=0$, SapBERT could be achieved best result on some cases like OET-disease.


When $d=0$, the task approximates near-equivalent matching. Under these conditions, SapBERT performs competitively, benefiting both from the properties of the OET datasets and its self-alignment training objective, which facilitates equivalent or near-equivalent matching. However, the MR scores highlight qualitative differences between SapBERT and the ontology embedding methods HiT and OnT. For instance, on EVAL-100, SapBERT, OnT, and HiT achieve the same MRR of 14, yet the MR disparity indicates that when SapBERT fails to find the optimal target early, its overall ranking is suboptimal. In contrast, HiT and OnT structurally encode concepts according to hierarchical depth. This encoding is advantageous because cosine similarity and lexical overlap alone are directionally agnostic, whereas HiT and OnT preserve hierarchical relationships. Consequently, concepts at greater depth than the query naturally appear toward the tail of the ranked list. This depth-sensitive ranking improves MR scores, a pattern observed on OET-CPP but less pronounced on OET-Disease, likely due to the higher proportion of near-equivalent query-target pairs.

Moreover, since $\lambda$ was tuned with a maximum depth threshold of 5, the closest, most specific subsumer may sometimes be overlooked in favour of distant, multi-hop targets. For example, OnT achieves an MRR of 15 and 16 ($d=1$, EVAL-100) for the miniature and full encoders, respectively, when ranking purely by hyperbolic distance (Appendix~\ref{appendix:full-experimental-results}). In such cases, ranking solely by $d_\kappa$ may outperform using $s(q, A)$ with $\lambda \gg 0$, particularly for genuine OOV queries. SapBERT appears more competitive on OET-CPP and preferable on OET-Disease because these evaluation sets contain queries favouring equivalence-based matching. Nevertheless, when both $s(q,A)$ and $d_\kappa$ are considered, HiT and OnT outperform all other baselines.

For multi-hop scenarios ($d \in \{2,4\}$), answer specificity decreases gradually, providing a more realistic measure of hierarchical retrieval than near-equivalent matching at $d=0$. In these settings, HiT and OnT consistently outperform baselines. From a practical standpoint, in applications like the SNOMED CT browser, an MR of 11--15 (HiT/OnT at $d=4$ on OET-CPP) ensures that a user entering an OOV query will encounter a valid ancestor on the first page of results. Conversely, an MR of 2898 (SapBERT) or 105,820 (TF-IDF) renders the system essentially unusable for OOV queries. For healthcare-adjacent applications, effective hierarchical retrieval can substantially improve conceptual recall, helping clinicians locate relevant concepts even when they do not know the exact terminology.

\section{Conclusion}
\label{sec:conclusion}

This work proposed an effective hierarchical retrieval framework for SNOMED CT on OOV queries, utilising language model-based ontology embeddings in hyperbolic space and a depth-based scoring function.
We found that the ontology embedding methods OnT and HiT perform similarly when applied in our HR framework. They consistently outperform all the lexical and language model-based baselines, and remain competitive with domain-tuned approaches such as SapBERT when the hierarchy structure is shallow, which resembles near-equivalent matching. However, as hierarchical structure and depth increases, becoming more relevant, both HiT and OnT demonstrate superior performance.
One limitation of this work is the annotation procedure for EVAL-100, which relies on a single individual. The evaluation dataset is also only 382 samples in total. This is partly due to manual annotation effort and low yields during sampling. However, it also points to an increasingly important area of research: the development of benchmarks for measuring structured knowledge retrieval.
Future work should continue to increase the size of the evaluation datasets, include domain-expert annotators, with an extension to ontologies beyond SNOMED CT, and investigate improvements to training strategies for effectively mapping ontologies at scale.

\bibliography{tgdk-v2021-sample-article}

\appendix

\clearpage

\section{Experimental Training Parameters (HiT, OnT)}
\label{appendix:full-experimental-settings}

\begin{table}[htbp]
    \centering
    \caption{Training configuraton and hyperparameter settings for HiT and OnT.}
    \label{tab:hit-ont-training-configuration}
    \begin{tabular}{lcc}
        \toprule
        \textbf{Parameter} & \textbf{HiT} & \textbf{OnT}             \\
        \midrule
        \multicolumn{3}{l}{\textit{Common, General}}                 \\ [2pt]
        Training epochs         & 4           & 4                    \\
        Training batch size     & 256         & 32                   \\
        Evaluation batch size   & 512         & 16                   \\
        Learning rate           & $1e{-5}$    & $1e{-5}$             \\
        \midrule
        \multicolumn{3}{l}{\textit{Contrastive Loss Functions}}      \\ [2pt]
        Clustering loss weight  & 1.0        & 1.0                   \\
        Clustering loss margin  & 3.0        & 3.0                   \\
        Centripetal loss weight & 1.0        & 1.0                   \\
        Centripetal loss margin & 0.5        & 0.5                   \\
        \midrule
        \multicolumn{3}{l}{\textit{Model-specific}}                  \\ [2pt]
        Negative sampling       & ``random'' & ---                   \\
        Role embedding mode     & ---        & ``sentenceEmbedding'' \\
        Role model mode         & ---        & ``rotation''          \\
        Existence loss kind     & ---        & ``hit''               \\
        Conjunction weight      & ---        & 1.0                   \\
        Existence weight        & ---        & 1.0                   \\
        \bottomrule
    \end{tabular}
\end{table}

\section{Appendix: Full Result Tables}
\label{appendix:full-experimental-results}

For HiT and OnT, besides their standard subsumption scores (defined in section~\ref{sec:hit-and-ont}), which combine hyperbolic distance and norms for an asymmetric, depth-biased ranking (favouring parent candidates with a hierarchical pre-order, as per \cite{yang2025language}), we also rank by hyperbolic distance directly, denoted $d_\kappa$. This is performed since the subsumption score includes a depth-bias term $\lambda$ that rewards candidates positioned higher (more general; towards the origin) in the hierarchy, which is useful for retrieving ancestors, but may overlook the most specific subsumers.

The specific SBERT PLM used for the reporting of these results is \textit{all-MiniLM-L12-v2}. All HiT and OnT models are trained in accordance with the configuration details provided under Appendix.~\ref{appendix:full-experimental-settings}. The results for EVAL-100 are shown in Table.~\ref{tab:eval-100-full}, the results for OET-CPP are shown in Table.~\ref{tab:OET-CPP-full}, and the results for OET-Disease are shown in Table.~\ref{tab:OET-DISEASE-full}.

\begin{table*}[htbp]
	\caption[OOV mention performance on EVAL-100]{Performance on EVAL-100 for $d \in \{1,3,5\}$.}
	\label{tab:eval-100-full}
	\footnotesize
	\centering
	\begin{tabular}{@{}ll ccc ccc ccc@{}}
		\toprule
                &             & \multicolumn{3}{c}{$d=0$} & \multicolumn{3}{c}{$d=2$} & \multicolumn{3}{c}{$d=4$} \\
                                \cmidrule(lr){3-5}          \cmidrule(lr){6-8}          \cmidrule(l){9-11}
		Method  & Metric      & MRR & H@$k$ & MR          & MRR & H@$k$ & MR          & MRR & H@$k$ & MR          \\
		\midrule
		TF-IDF  & ---         & 04 & 02/03/04 & 251820    & 04 & 02/04/05 & 244291    & 05 & 02/04/06 & 244290    \\
		BM25    & ---         & 06 & 05/06/07 & 101885    & 07 & 05/07/08 & 75377     & 07 & 05/07/08 & 64054     \\
		SBERT   & Cosine sim. & 11 & 05/14/16 & 20127     & 12 & 05/15/17 & 11994     & 12 & 05/15/17 & 10699     \\
		SapBERT & Cosine sim. & 14 & 09/14/20 & 26942     & 15 & 09/15/20 & 19456     & 15 & 09/15/20 & 14997     \\
		\midrule
		HiT (M) & $d_k$       & 11 & 06/10/17 & 4171      & 15 & 09/13/22 & 2002      & 15 & 09/13/22 & 1736      \\
		HiT (M) & $s(q,A)$    & 06 & 05/05/06 & 3773      & 11 & 08/10/13 & 1163      & 16 & 11/14/20 & 733       \\
		HiT (F) & $d_k$       & 13 & 07/17/18 & 2771      & 16 & 09/20/23 & 986       & 17 & 10/20/23 & 705       \\
		HiT (F) & $s(q,A)$    & 14 & 09/15/17 & 2647      & 22 & 15/22/30 & 722       & 24 & 16/25/33 & 392       \\
		\midrule
		OnT (M) & $d_k$       & 15 & 09/17/22 & 3787      & 18 & 11/21/26 & 1194      & 19 & 12/22/27 & 875       \\
		OnT (M) & $s(q,A)$    & 14 & 10/15/17 & 2494      & 18 & 12/19/22 & 577       & 20 & 13/22/26 & 303       \\
		OnT (F) & $d_k$       & 16 & 12/17/21 & 3312      & 18 & 12/19/25 & 1512      & 19 & 13/19/25 & 1044      \\
		OnT (F) & $s(q,A)$    & 13 & 06/16/19 & 2179      & 16 & 06/20/26 & 646       & 18 & 06/21/29 & 317       \\
		\bottomrule
	\end{tabular}
\end{table*}

\begin{table*}[htbp]
	\caption[OOV mention performance on OET-CPP]{Performance on OET-CPP for $d \in \{1,3,5\}$.}
	\label{tab:OET-CPP-full}
	\footnotesize
	\centering
	\begin{tabular}{@{}ll ccc ccc ccc@{}}
		\toprule
                &             & \multicolumn{3}{c}{$d=0$} & \multicolumn{3}{c}{$d=2$} & \multicolumn{3}{c}{$d=4$} \\
                                \cmidrule(lr){3-5}          \cmidrule(lr){6-8}          \cmidrule(l){9-11}
		Method  & Metric      & MRR & H@$k$ & MR          & MRR & H@$k$ & MR          & MRR & H@$k$ & MR          \\
		\midrule
		TF-IDF  & ---         & 09 & 07/10/11 & 107927    & 10 & 07/12/13 & 105832    & 10 & 07/12/13 & 105820    \\
		BM25    & ---         & 10 & 08/11/11 & 51081     & 11 & 09/12/13 & 42459     & 12 & 10/13/13 & 41353     \\
		SBERT   & Cosine sim. & 17 & 10/20/24 & 12683     & 19 & 11/23/28 & 3923      & 19 & 11/23/28 & 3129      \\
		SapBERT & Cosine sim. & 25 & 17/28/34 & 6667      & 26 & 18/29/36 & 3552      & 26 & 18/29/36 & 2898      \\
		\midrule
		HiT (M) & $d_\kappa$  & 22 & 12/25/32 & 1081      & 26 & 16/29/35 & 128       & 27 & 16/31/37 & 68        \\
		HiT (M) & $s(q,A)$    & 17 & 10/18/22 & 1292      & 38 & 25/45/52 & 52        & 49 & 33/59/66 & 12        \\
		HiT (F) & $d_\kappa$  & 21 & 13/22/32 & 1543      & 26 & 17/28/37 & 91        & 27 & 17/29/37 & 60        \\
		HiT (F) & $s(q,A)$    & 11 & 06/11/14 & 1827      & 33 & 19/41/50 & 55        & 54 & 41/62/70 & 11        \\
		\midrule
		OnT (M) & $d_\kappa$  & 23 & 14/28/33 & 1313      & 27 & 17/32/40 & 1120      & 29 & 17/35/41 & 71        \\
		OnT (M) & $s(q,A)$    & 25 & 14/29/38 & 1339      & 38 & 24/46/54 & 46        & 42 & 26/49/59 & 14        \\
		OnT (F) & $d_\kappa$  & 20 & 11/23/31 & 1272      & 23 & 14/26/35 & 149       & 25 & 14/27/38 & 102       \\
		OnT (F) & $s(q,A)$    & 25 & 17/28/31 & 1170      & 39 & 27/43/49 & 51        & 47 & 32/54/63 & 15        \\
		\bottomrule
	\end{tabular}
\end{table*}

\begin{table*}[htbp]
	\caption[OOV mention performance on OET-Disease]{Performance on OET-Disease for $d \in \{1,3,5\}$.}
	\label{tab:OET-DISEASE-full}
	\footnotesize
	\centering
	\begin{tabular}{@{}ll ccc ccc ccc@{}}
		\toprule
                &             & \multicolumn{3}{c}{$d=0$} & \multicolumn{3}{c}{$d=2$} & \multicolumn{3}{c}{$d=4$} \\
                                \cmidrule(lr){3-5}          \cmidrule(lr){6-8}          \cmidrule(l){9-11}
		Method  & Metric      & MRR & H@$k$ & MR          & MRR & H@$k$ & MR          & MRR & H@$k$ & MR          \\
		\midrule
		TF-IDF  & ---         & 12 & 07/15/16 & 82027     & 13 & 07/16/18 & 75686     & 13 & 07/16/18 & 75651     \\
		BM25    & ---         & 12 & 09/11/13 & 30056     & 14 & 11/13/15 & 23499     & 14 & 11/13/15 & 21960     \\
		SBERT   & Cosine sim. & 23 & 16/29/31 & 3804      & 28 & 18/36/38 & 1282      & 28 & 18/36/38 & 793       \\
		SapBERT & Cosine sim. & 38 & 27/40/49 & 1300      & 39 & 27/42/55 & 1089      & 39 & 27/42/55 & 968       \\
		\midrule
		HiT (M) & $d_\kappa$  & 25 & 13/29/36 & 2530      & 31 & 16/40/45 & 124       & 31 & 16/40/45 & 56        \\
		HiT (M) & $s(q,A)$    & 20 & 15/24/24 & 3057      & 36 & 24/47/53 & 59        & 50 & 35/62/65 & 11        \\
		HiT (F) & $d_\kappa$  & 27 & 16/29/40 & 3552      & 33 & 22/36/44 & 89        & 35 & 24/38/45 & 56        \\
		HiT (F) & $s(q,A)$    & 11 & 07/13/15 & 3855      & 20 & 11/22/27 & 88        & 46 & 36/49/56 & 17        \\
		\midrule
		OnT (M) & $d_\kappa$  & 28 & 15/38/42 & 3262      & 33 & 18/45/49 & 144       & 34 & 18/47/51 & 75        \\
		OnT (M) & $s(q,A)$    & 30 & 16/40/51 & 3443      & 44 & 25/56/65 & 60        & 46 & 25/60/71 & 11        \\
		OnT (F) & $d_\kappa$  & 26 & 15/31/38 & 2684      & 29 & 18/35/44 & 167       & 32 & 20/36/47 & 123       \\
		OnT (F) & $s(q,A)$    & 31 & 22/38/40 & 2561      & 41 & 29/47/51 & 67        & 47 & 31/53/65 & 23        \\
		\bottomrule
	\end{tabular}
\end{table*}

\end{document}